\documentclass[conference]{IEEEtran}
\IEEEoverridecommandlockouts
\usepackage{cite}
\usepackage{amsmath,amssymb,amsfonts}
\usepackage{algorithmic}
\usepackage{graphicx}
\usepackage{textcomp}
\usepackage{xcolor}

\usepackage{booktabs}
\usepackage{multirow}
\usepackage{url}
\usepackage{hyperref}
\usepackage{extra}
\usepackage{caption}
\usepackage{subcaption}

\hypersetup{
    colorlinks=true,
    linkcolor=blue,
    filecolor=magenta,      
    urlcolor=cyan,
    citecolor=blue,
    pdfpagemode=FullScreen,
    }

\def\BibTeX{{\rm B\kern-.05em{\sc i\kern-.025em b}\kern-.08em
    T\kern-.1667em\lower.7ex\hbox{E}\kern-.125emX}}
\begin{document}

\title{SID: Stereo Image Dataset for Autonomous Driving in Adverse Conditions\\
\thanks{This project was supported by a research initiation and development grant from the office of research and sponsored programs at the University of Michigan-Dearborn.}
}

\author{\IEEEauthorblockN{\small Zaid A. El-Shair, Abdalmalek Abu-raddaha, Aaron Cofield, Hisham Alawneh, Mohamed Aladem, Yazan Hamzeh, Samir A. Rawashdeh}
\IEEEauthorblockA{\textit{Electrical and Computer Engineering Department} \\
\textit{University of Michigan-Dearborn}\\
Dearborn, MI, USA \\
\{zelshair, abdmalek, afcofiel, haalawne, maladem, yhamzeh, srawa\}@umich.edu}}

\maketitle

\begin{abstract}
Robust perception is critical for autonomous driving, especially under adverse weather and lighting conditions that commonly occur in real-world environments. In this paper, we introduce the Stereo Image Dataset (SID), a large-scale stereo-image dataset that captures a wide spectrum of challenging real-world environmental scenarios. Recorded at a rate of 20 Hz using a ZED stereo camera mounted on a vehicle, SID consists of 27 sequences totaling over 178k stereo image pairs that showcase conditions from clear skies to heavy snow, captured during the day, dusk, and night.
The dataset includes detailed sequence-level annotations for weather conditions, time of day, location, and road conditions, along with instances of camera lens soiling, offering a realistic representation of the challenges in autonomous navigation.
Our work aims to address a notable gap in research for autonomous driving systems by presenting high-fidelity stereo images essential for the development and testing of advanced perception algorithms. These algorithms support consistent and reliable operation across variable weather and lighting conditions, even when handling challenging situations like lens soiling. SID is publicly available at: \url{https://doi.org/10.7302/esz6-nv83}.

\end{abstract}


\section{Introduction}

Autonomous vehicles (AVs) and Advanced Driver-Assistance Systems (ADAS) rely heavily on robust perception capabilities to navigate and interact with their environment safely and effectively \cite{CGV-079, s19030648}. Central to these capabilities is the quality and diversity of the data upon which perception algorithms are trained. Currently, there is a critical challenge faced by autonomous systems—achieving reliable and consistent performance under varying conditions and locations that a vehicle may encounter during real-world operation \cite{CGV-079}. Moreover, the majority of datasets available today do not fully account for the complexities of adverse weather conditions or the intricate lighting variations that span from dawn to dusk.
Hence, the motivation behind new dataset development is twofold. Firstly, there is a need for models that maintain their accuracy and reliability in diverse weather scenarios, from clear skies to heavy snow. Such models are crucial for AVs that must operate across different geographies and seasons \cite{8500543}. Secondly, the deployment of AVs and ADAS in diverse settings demands a dataset that covers a wide range of urban and non-urban environments \cite{s22218493,chen2017multi}. This ensures that automotive systems are attuned to the nuances of various surroundings, from the structured layouts of campuses to the unpredictable dynamics of residential areas. 
Additionally, in the field of perception, stereo vision stands as a critical component for depth estimation---a vital process that enables AVs to discern distance and spatial relationships between objects. Stereo vision algorithms simulate human binocular vision by utilizing pairs of images, each taken from slightly different viewpoints \cite{aladem2019comparative}. By comparing these images, the algorithms can detect disparities that are translated into depth information. This depth perception is fundamental for autonomous vehicles to identify potential obstacles, estimate distances, and make informed navigation decisions. However, the reliability of depth estimation can be significantly impacted by adverse weather conditions that lead to visual distortions or camera lens soiling, which are often underrepresented in existing datasets.

To address these challenges, we introduce the Stereo Image Dataset (SID) for autonomous driving in adverse conditions. Recorded using a ZED stereo camera mounted on a vehicle, SID includes 27 sequences that capture a variety of weather conditions and lighting scenarios, ranging from clear daylight to challenging snow sequences at night.
This dataset is designed to support the development and evaluation of perception algorithms with high-resolution stereo images, particularly for conditions that have been underrepresented in existing datasets, such as snow and rain. SID aims to contribute to advancements in robust autonomous driving technologies. This dataset has been deposited in Deep Blue Data and is publicly accessible at \cite{SID_data_deposit}.



\begin{figure}[t]
     \centering
     \begin{subfigure}[b]{0.24\textwidth}
         \centering
         \frame{\includegraphics[width=\textwidth]{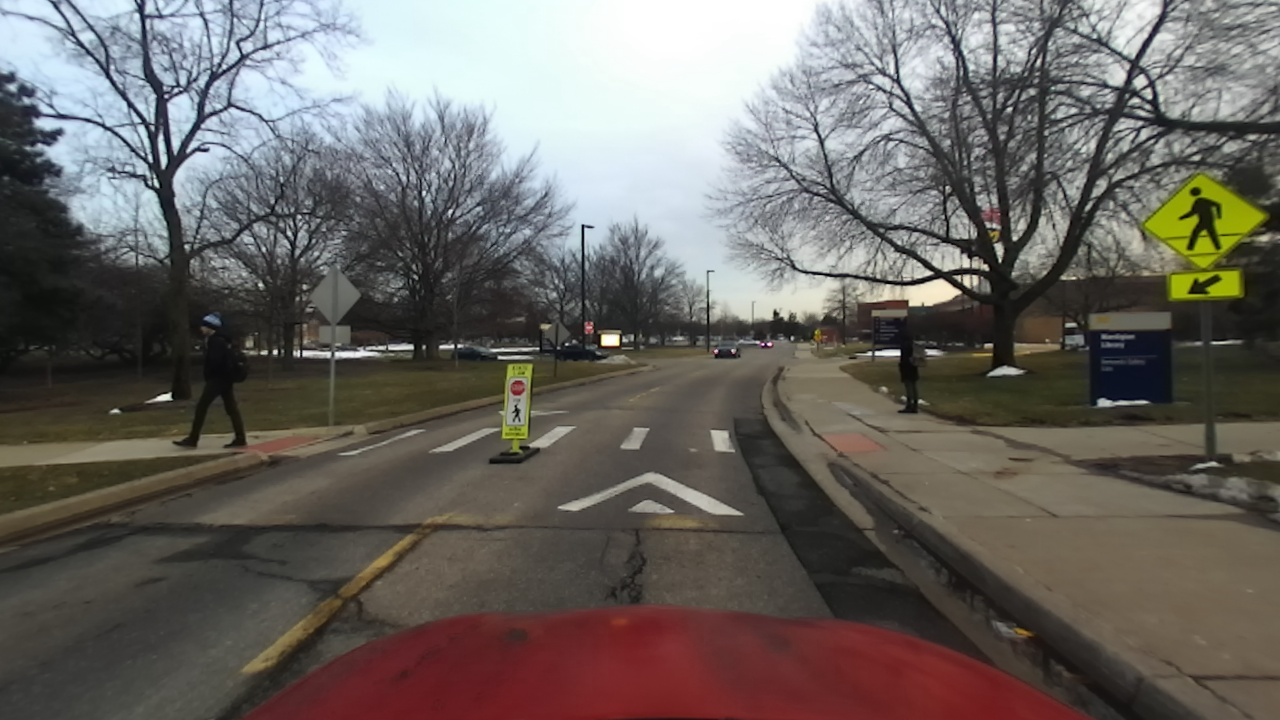}}
     \end{subfigure}
     \begin{subfigure}[b]{0.24\textwidth}
         \centering
         \frame{\includegraphics[width=\textwidth]{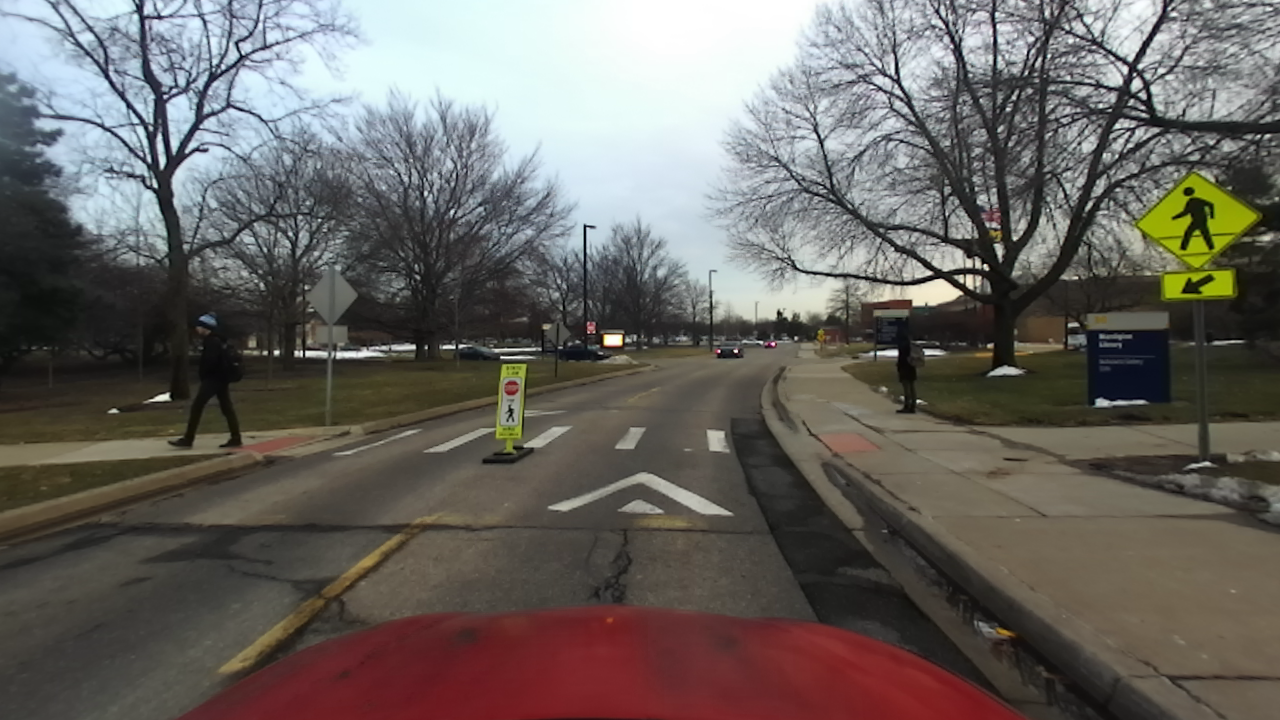}}
     \end{subfigure}
      \caption{Stereo image sample showing the left and right images captured by the ZED camera at 720p resolution.}
      \label{fig:sample}
\end{figure}

\section{Related Work}

The advancements of perception systems in AVs and ADAS have seen considerable contributions from datasets like KITTI \cite{geiger2012we}, Cityscapes \cite{cordts2016cityscapes}, and nuScenes \cite{caesar2020nuscenes}. These datasets provide benchmarks for a range of computer vision tasks predominantly under moderate weather conditions and standard urban scenarios. Large-scale datasets such as BDD100K \cite{yu2020bdd100k}, with its variety of weather scenes, and Mapillary Vistas \cite{neuhold2017mapillary}, with its broad geographical settings, have expanded environmental diversity. However, they do not offer the stereo image pairs which are essential for accurate depth perception in AV and ADAS perception systems. Further, they are limited to short video durations \cite{yu2020bdd100k} and single-frame images \cite{neuhold2017mapillary}.



Emerging datasets have introduced specialized scenarios that present their unique sets of challenges. Synthia \cite{ros2016synthia}, with its synthetic generation of varied weathers, provides annotated scenes for segmentation algorithms under different weather elements. The Oxford RobotCar dataset \cite{maddern20171} presents an impressive range of data collected over an extended period, highlighting changes in the environment across seasons. RADIATE \cite{sheeny2020radiate} focuses on the often-overlooked yet critical scenario of radar data in adverse weather conditions. The expansive Waymo Open Dataset \cite{sun2020scalability} incorporates a multi-sensor array that includes high-quality imagery and LIDAR data under various weather and lighting conditions. Although these datasets greatly enhance the field, they often lack detailed representation across adverse weather and lighting conditions necessary for thorough algorithmic training and testing.

SID seeks to bridge these critical gaps in AV and ADAS research by providing a comprehensive stereo-image dataset that covers a wide range of environmental conditions encountered in real-world scenarios. This dataset comprises 27 high-resolution stereo-image sequences, each averaging around 6 minutes in length. These sequences feature five distinct weather conditions from clear to heavy snow, three different times of the day (day, dusk, and night), and various road conditions captured across multiple locations. In addition to its environmental diversity, SID captures operational challenges like camera lens soiling during heavy snowfall, offering researchers the opportunity to develop navigation systems robust against such occurrences. The dataset is split into training and testing sets to assist in the development and evaluation of perception models. With its richly annotated sequences and a focus on temporal and spatial details, SID offers valuable data for designing perception algorithms that enhance the safety and dependability of AV and ADAS navigation.

\begin{figure}[t]
    \centering
    \begin{subfigure}[b]{0.24\textwidth}
        \frame{\includegraphics[width=\textwidth]{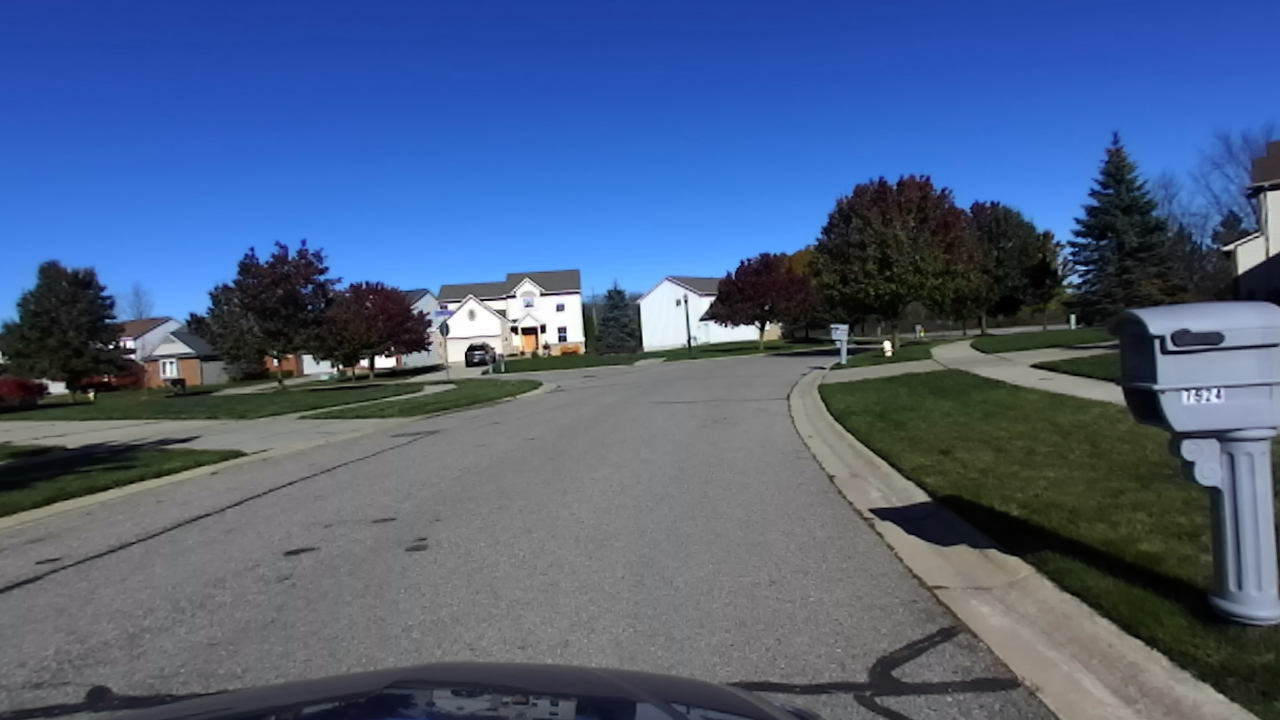}}
        \caption{}
    \end{subfigure}
    \begin{subfigure}[b]{0.24\textwidth}
        \frame{\includegraphics[width=\textwidth]{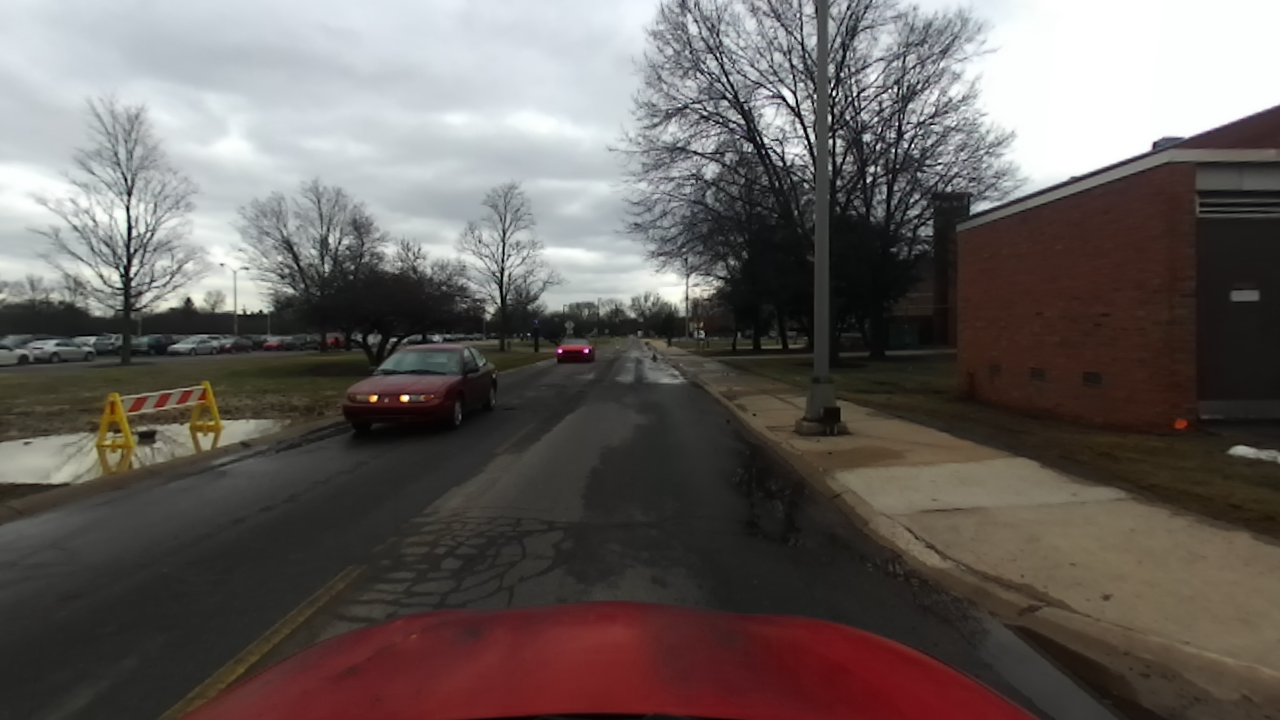}}
        \caption{}
    \end{subfigure}

    \smallskip
    
    \begin{subfigure}[b]{0.24\textwidth}
        \frame{\includegraphics[width=\textwidth]{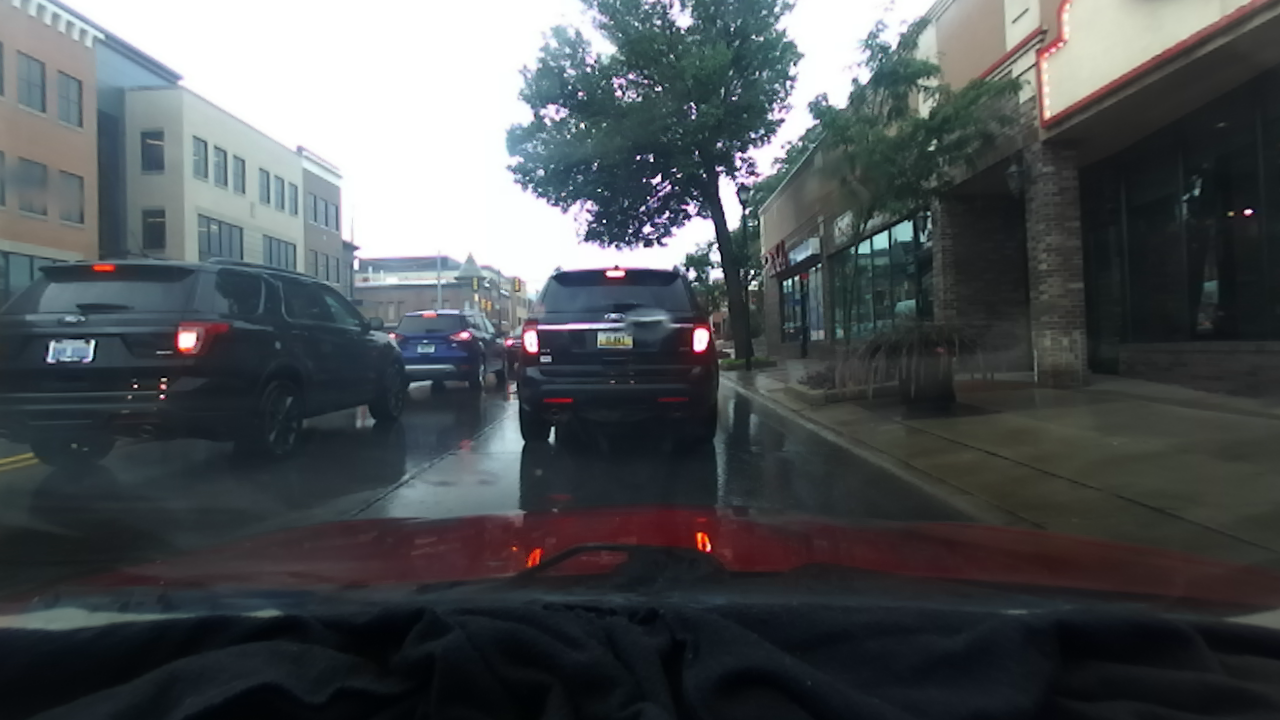}}
        \caption{}
    \end{subfigure}
    \begin{subfigure}[b]{0.24\textwidth}
        \frame{\includegraphics[width=\textwidth]{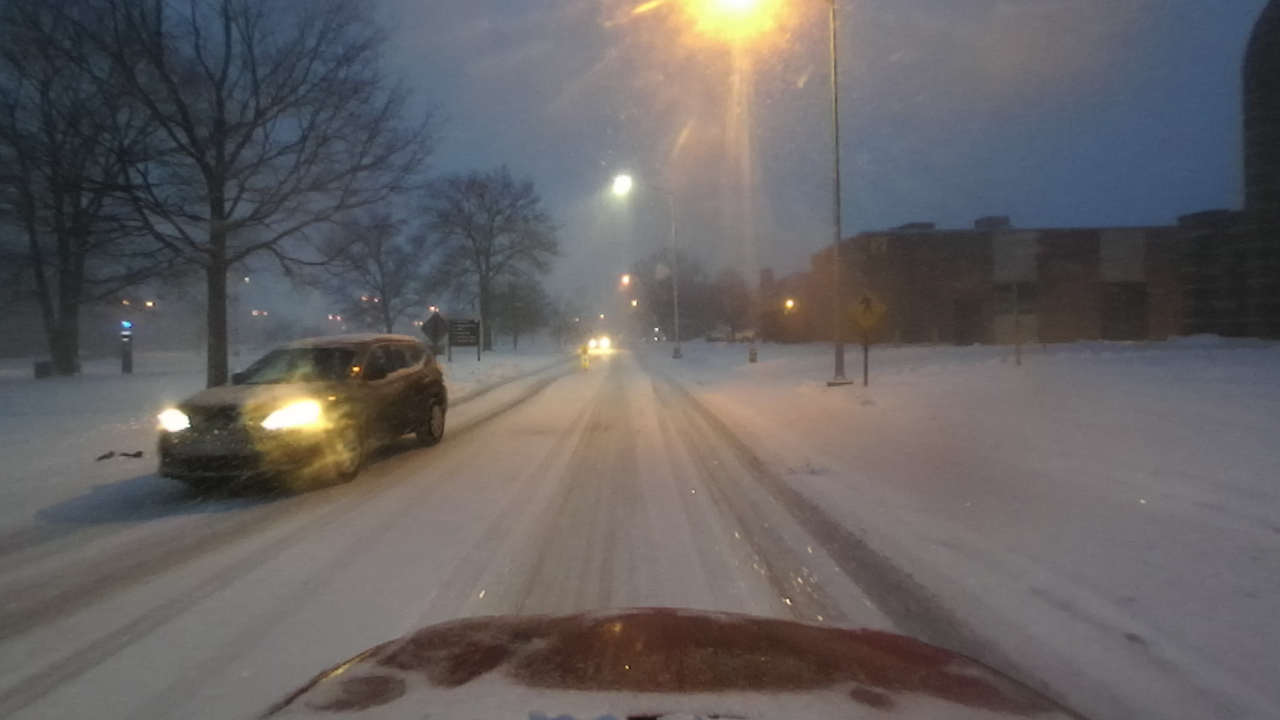}}
        \caption{}
    \end{subfigure}    
    \caption{Multiple weather conditions included in our dataset such as (a) \textit{Clear}, (b) \textit{Overcast}, (c) \textit{Rain}, and (d) \textit{Snow}.}
    \label{fig:weather_conditions}
\end{figure}

\begin{figure}
    \centering
    \begin{subfigure}[b]{0.24\textwidth}
        \frame{\includegraphics[width=\textwidth]{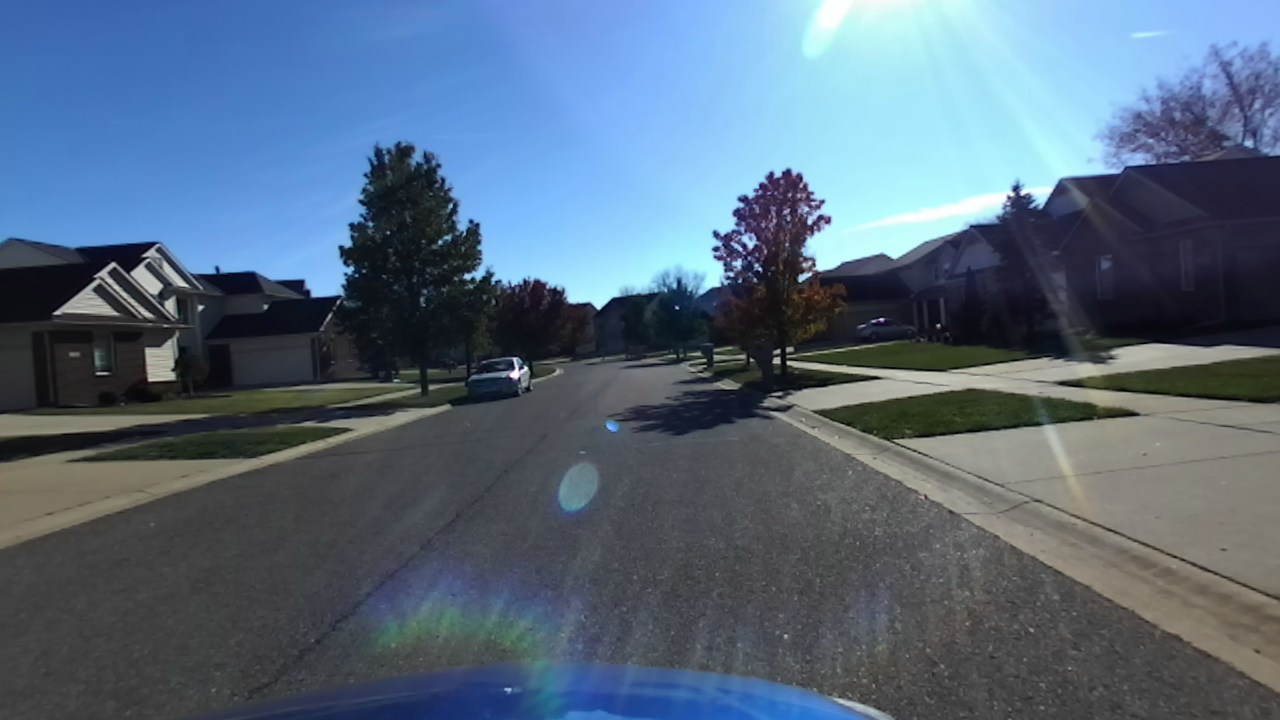}}
        \caption{}
    \end{subfigure}
    \begin{subfigure}[b]{0.24\textwidth}
        \frame{\includegraphics[width=\textwidth]{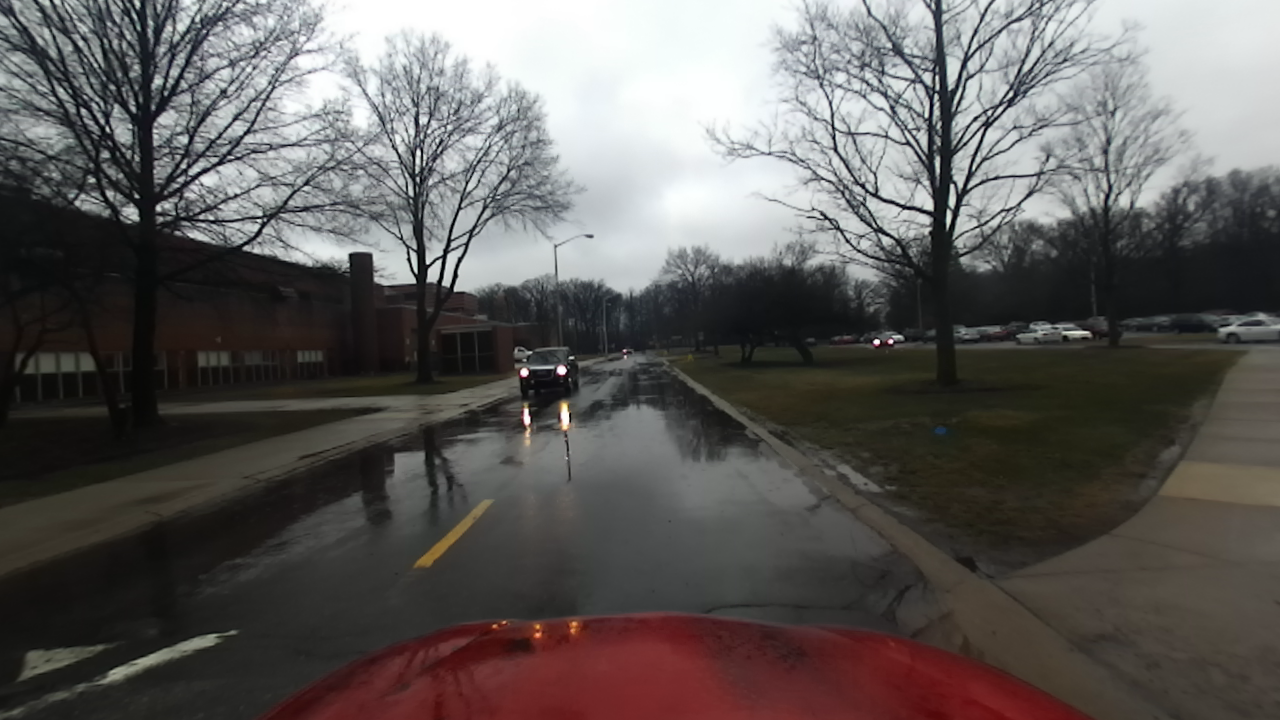}}
        \caption{}
    \end{subfigure}

    \smallskip
    
    \begin{subfigure}[b]{0.24\textwidth}
        \frame{\includegraphics[width=\textwidth]{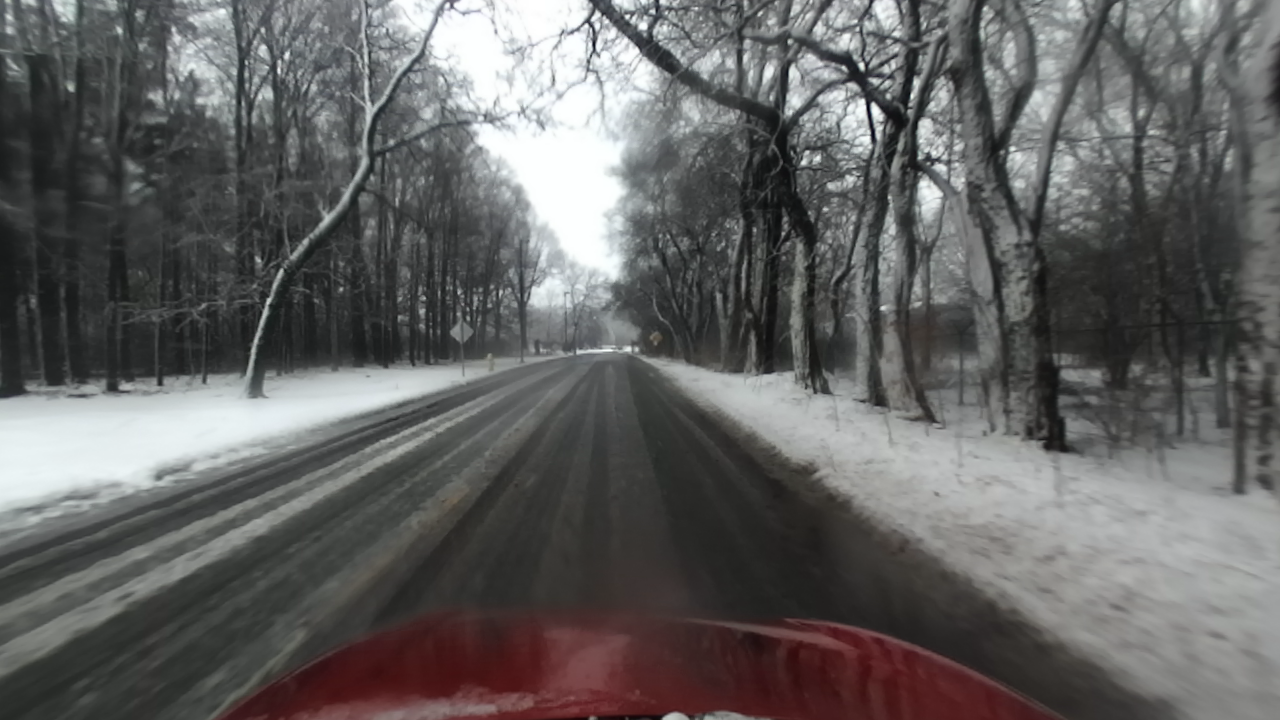}}
        \caption{}
    \end{subfigure}
    \begin{subfigure}[b]{0.24\textwidth}
        \frame{\includegraphics[width=\textwidth]{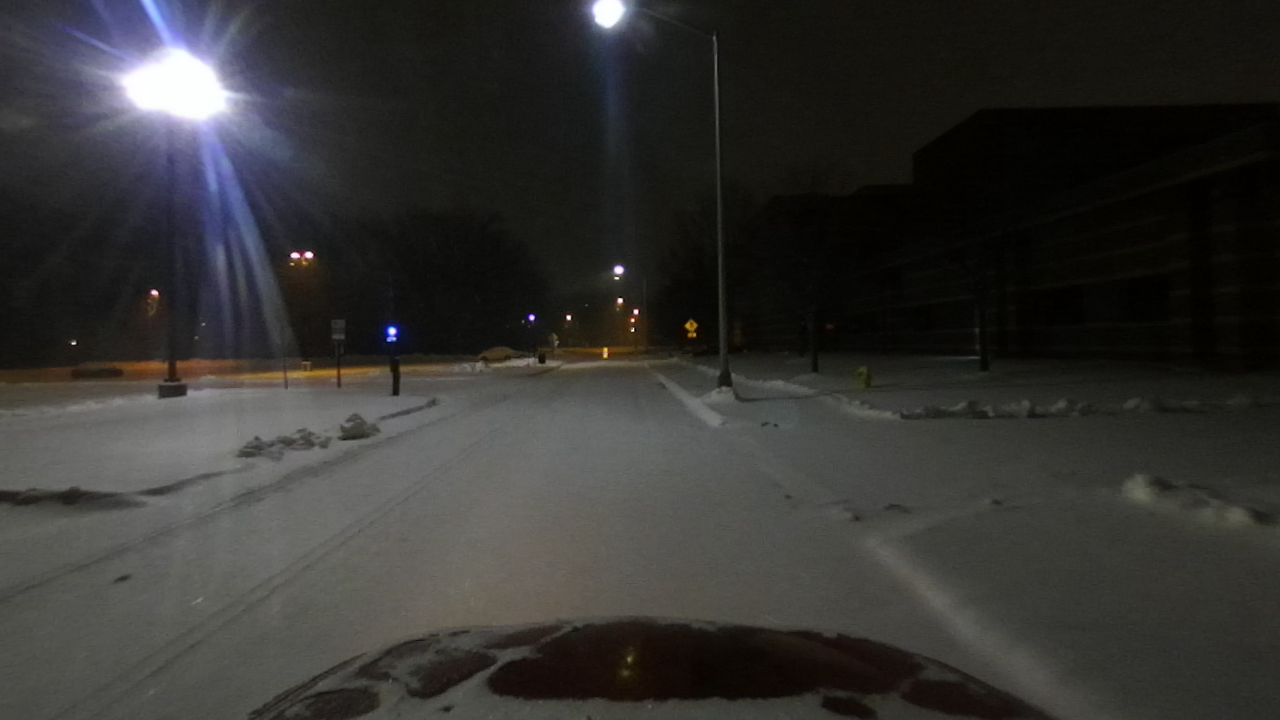}}
        \caption{}
    \end{subfigure}    
    \caption{Various road conditions present in our dataset such as (a) \textit{Dry}, (b) \textit{Wet}, (c) \textit{Snow and Wet}, and (d) \textit{Snowy}.}
    \label{fig:road_conditions}
\end{figure}

\begin{figure}[!h]
    \centering
    \begin{subfigure}{0.24\textwidth}
        \frame{\includegraphics[width=\textwidth]{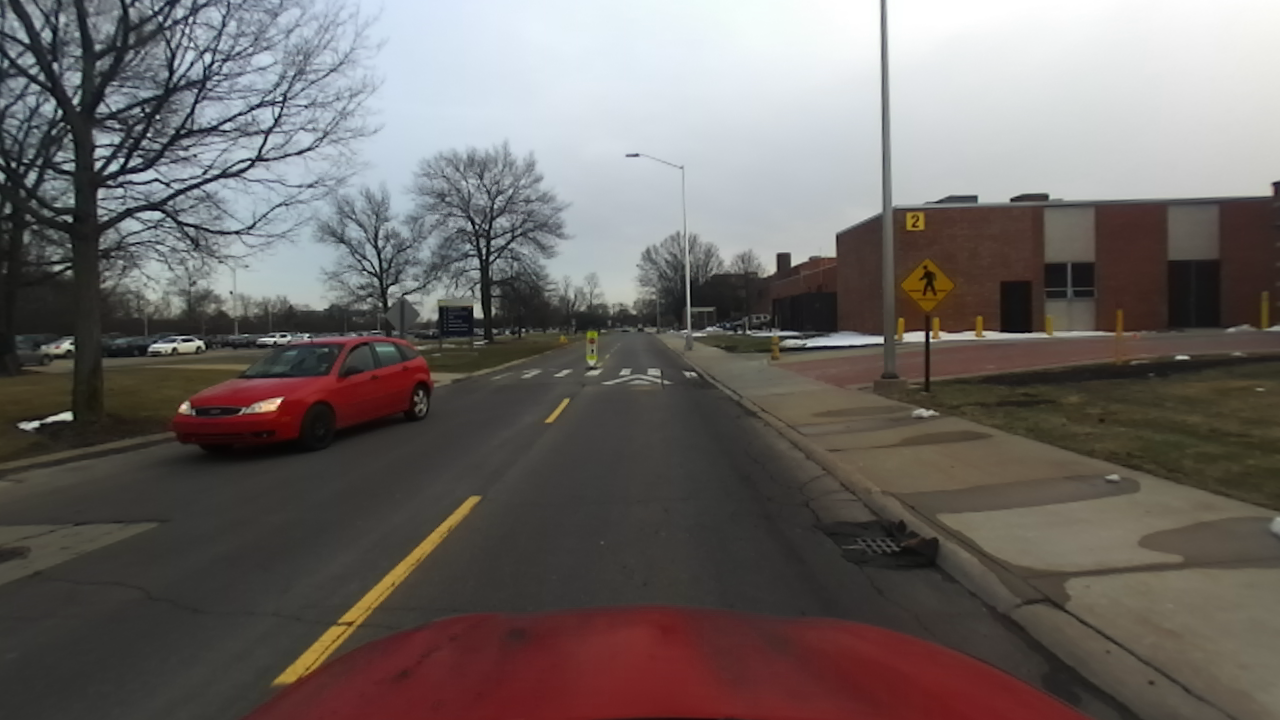}}
        \caption{}
    \end{subfigure}
    \begin{subfigure}{0.24\textwidth}
        \frame{\includegraphics[width=\textwidth]{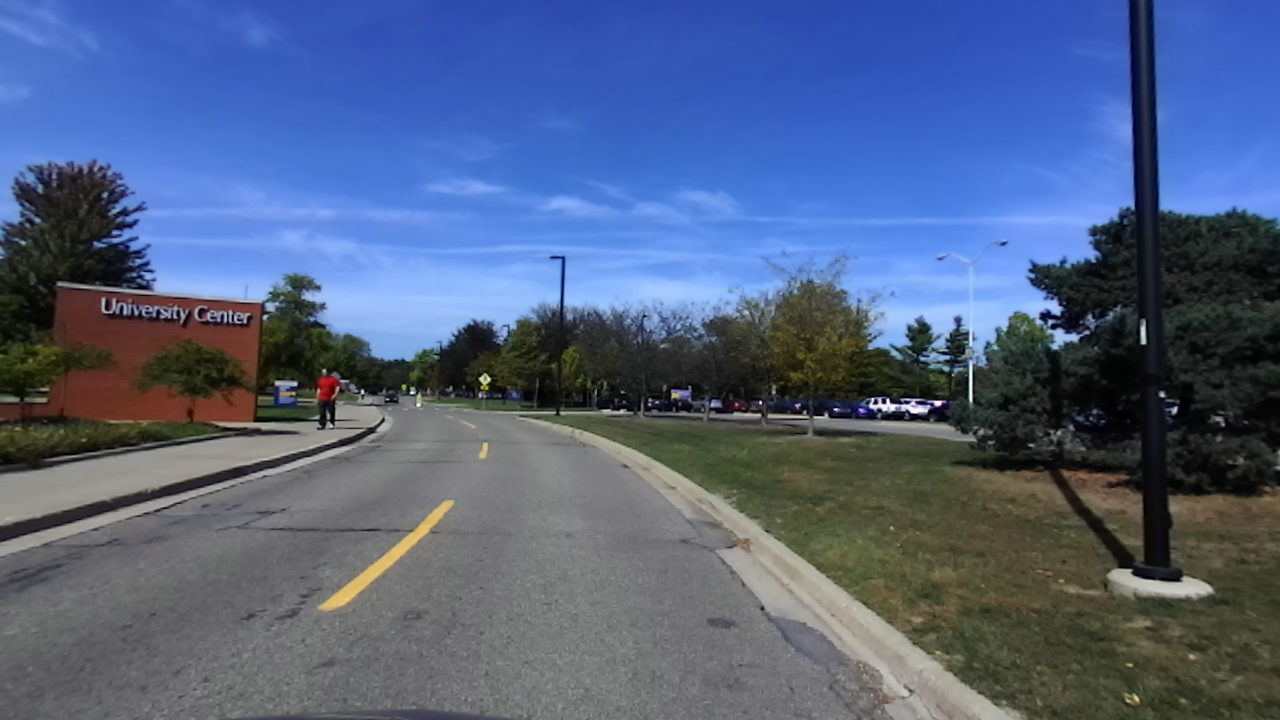}}
        \caption{}
    \end{subfigure}
    
    \smallskip
    \begin{subfigure}{0.24\textwidth}
        \frame{\includegraphics[width=\textwidth]{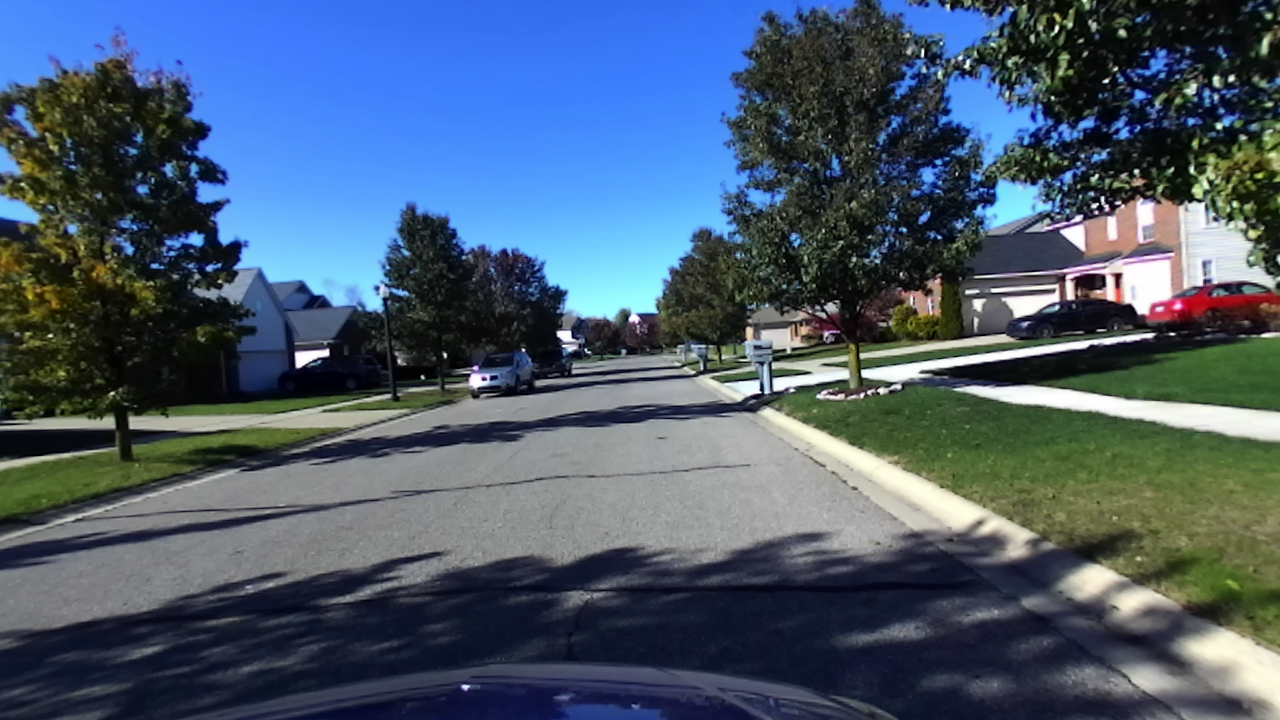}}
        \caption{}
    \end{subfigure}
    \begin{subfigure}{0.24\textwidth}
        \frame{\includegraphics[width=\textwidth]{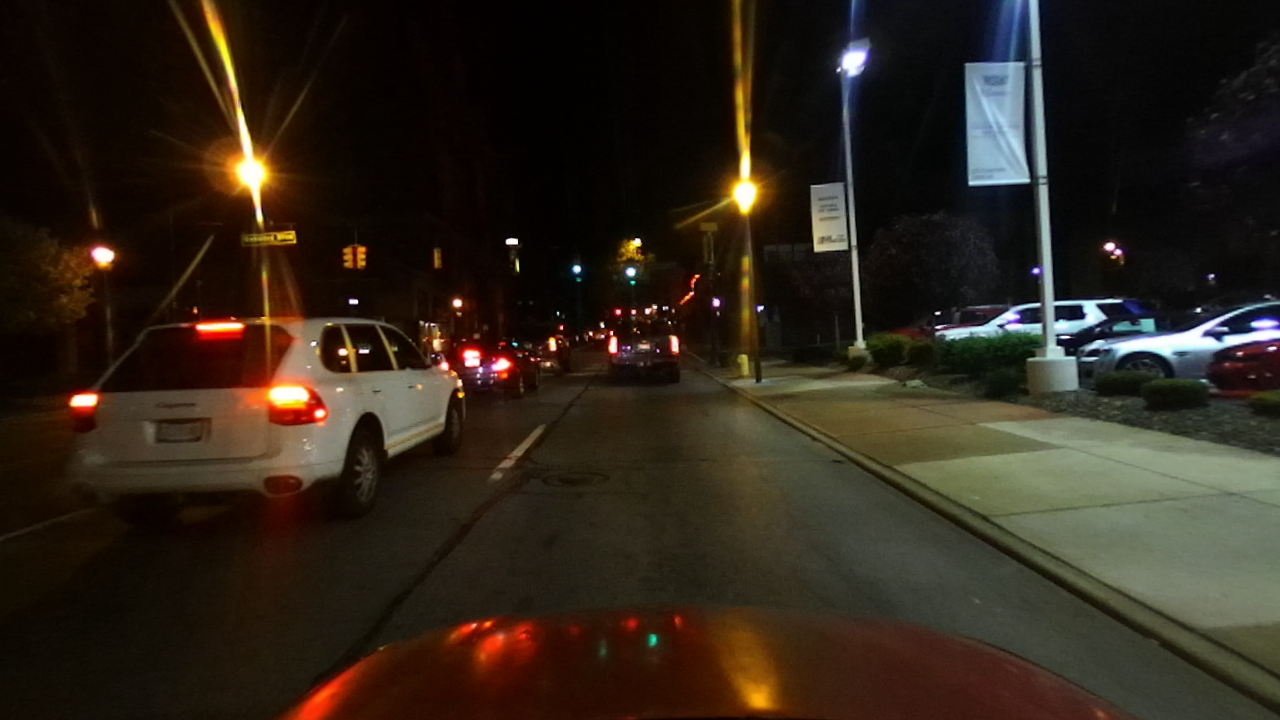}}
        \caption{}
    \end{subfigure}    
    \caption{Locations available in our dataset including (a) \textit{Campus (CW)}, (b) \textit{Campus (CCW)}, (c) \textit{Residential}, and (d) \textit{City}.}
    \label{fig:locations}
\end{figure}

\section{Data Collection}

\subsection{Data Collection Methodology and Hardware}
SID was recorded using the ZED camera, a commercially available stereo-vision camera produced by StereoLabs \cite{stereolabsStereolabsPerception}. This camera can capture stereo high-definition images providing the ability to estimate depth which is crucial for various perception tasks in robotics. In the majority of the recorded sequences, the camera was mounted on top of the moving data collection vehicle to capture stereo image sequences. However, in a few sequences under heavy rain, the camera was positioned below the windshield to shield it from the elements and facilitate data capture.

The ZED camera interfaced with a laptop equipped with an Nvidia graphics processing unit (GPU)—a requirement for the camera's application programming interface (API)—an Intel i5-7200 central processing unit (CPU), 8GB of random access memory (RAM), and a 256GB solid-state drive (SSD), operating on Ubuntu 16.04. We developed a custom data logging software employing the ZED camera API, which supports various resolutions such as VGA, HD 720p, HD 1080p, and 2K. Opting for a 720p (1280$\times$720 pixels) resolution allowed us to capture image sequences at 20 Hz without compromising the quality necessary for depth perception. The camera has been factory-calibrated, with the API providing rectified images. We initially stored the rectified images as bitmap (.bmp) files before converting them to the highly compressed but lossless Portable Network Graphics (.png) format to optimize storage efficiency.

\subsection{Environmental Conditions and Locations}

The data collection spanned an entire year to ensure the inclusion of a range of weather conditions and varying lighting environments. The weather parameters for the dataset encompass Clear, Cloudy, Overcast, Rain, and Snow conditions, as illustrated in Fig. \ref{fig:weather_conditions}. To capture a variety of lighting scenarios, we recorded image sequences during the Day, Night, and a few at Dusk. These varying times of day and weather patterns resulted in diverse road conditions including Snowy, Wet, and Dry, among others as shown in Fig. \ref{fig:road_conditions}.

Multiple recording sites were chosen to enhance the dataset's richness. Predominantly, sequences were captured within the grounds of the University of Michigan-Dearborn campus. The campus terrain is favored due to consistent traffic flow and diverse structural environments. Our defined route on campus presents inconsistent lane markings, zones with and without curbsides, regions dominated by artificial structures like buildings and signage, as well as areas abundant with natural features such as trees and greenery. It also includes shaded sections, open spaces, and places that encounter both pedestrian and vehicle traffic. Other recording locations encompassed residential neighborhoods and various urban settings, which added to the heterogeneity of the dataset. The selected locations are shown in Fig. \ref{fig:locations}




\begin{figure*}[t]
     \begin{subfigure}[b]{0.33\textwidth}
         \centering
         \includegraphics[width=\textwidth]{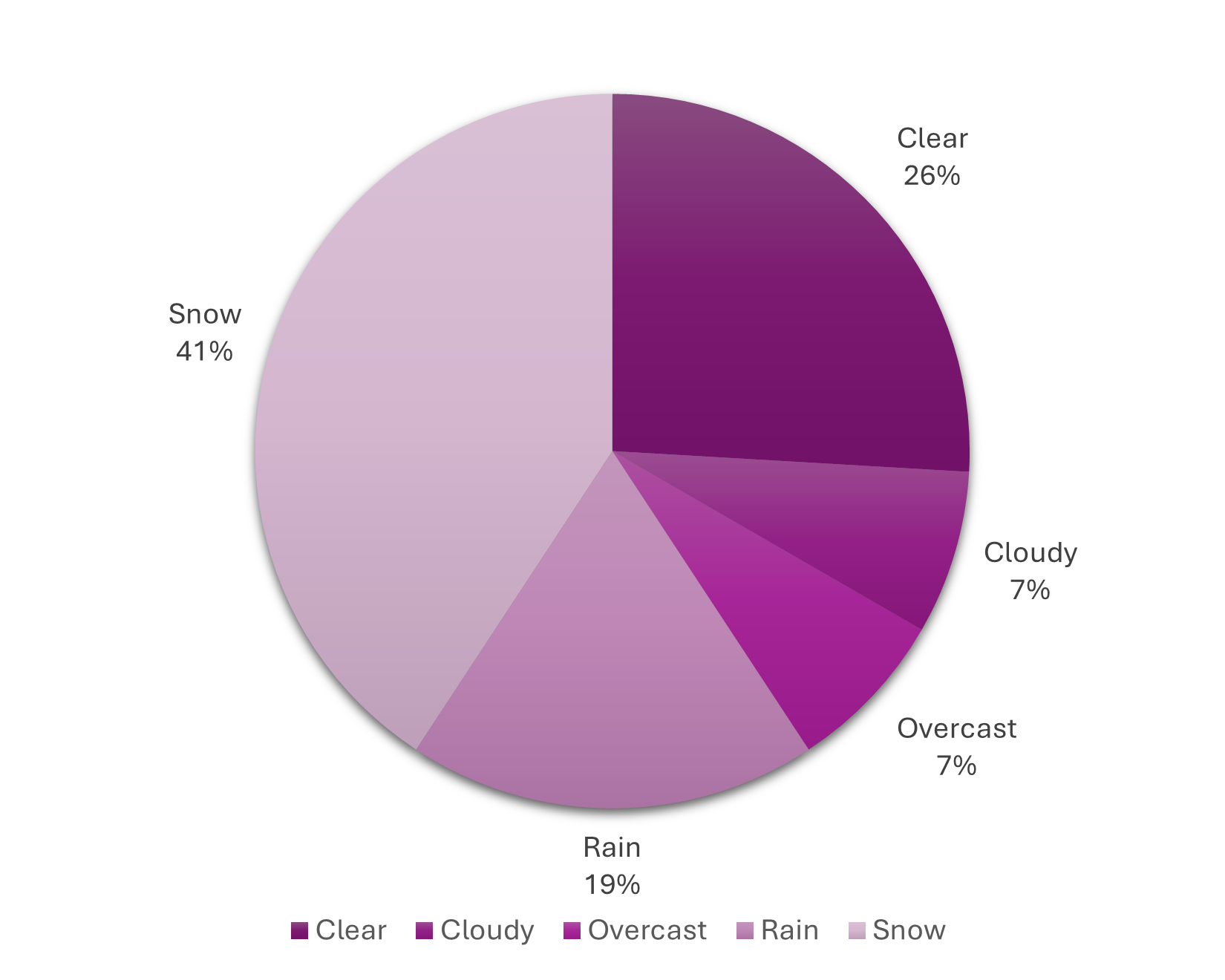}
         \caption{Weather Condition}
         \label{fig:pie_a}
     \end{subfigure}
     \begin{subfigure}[b]{0.33\textwidth}
         \centering
         \includegraphics[width=\textwidth]{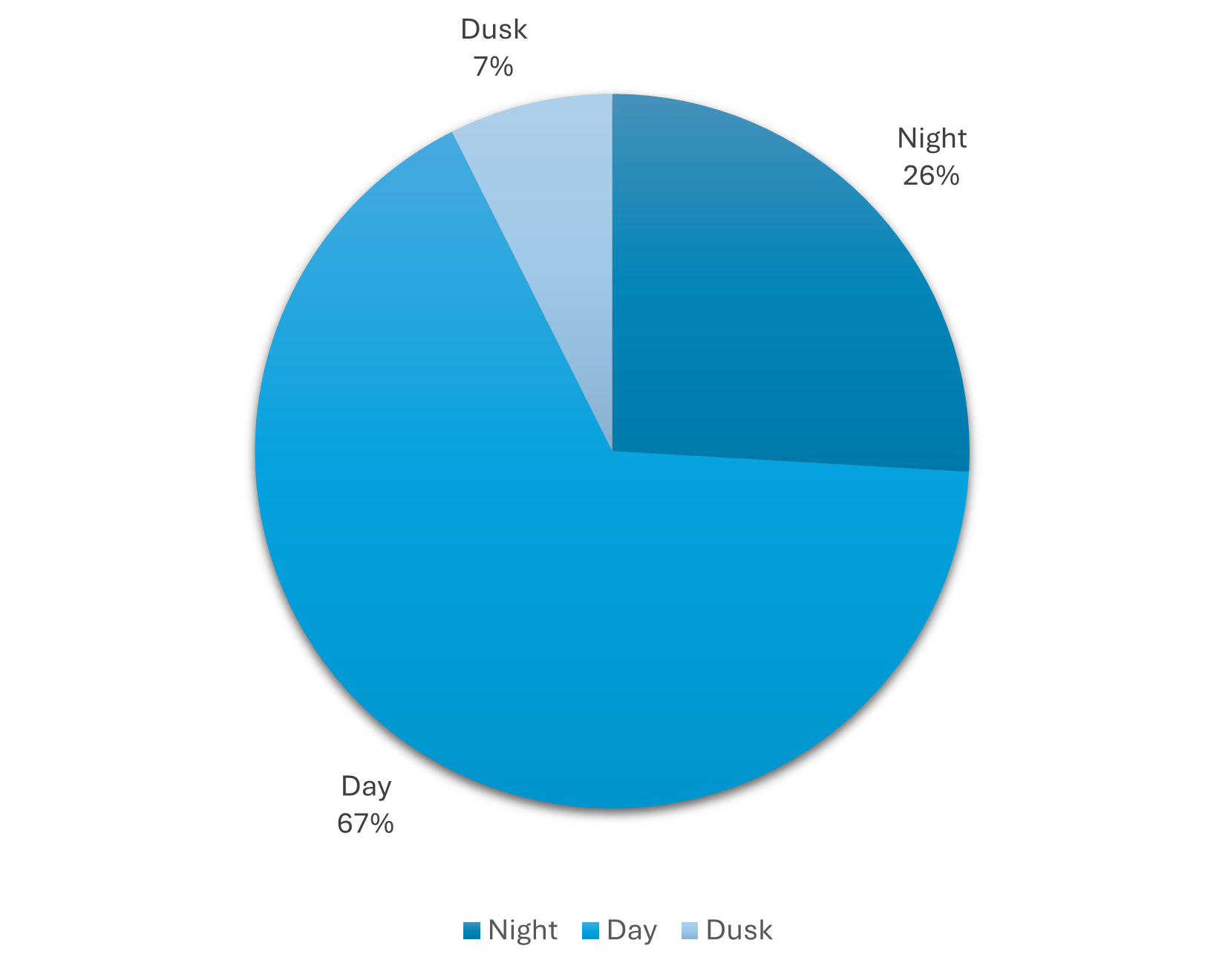}
         \caption{Time of Day}
         \label{fig:pie_b}
     \end{subfigure}
     \begin{subfigure}[b]{0.33\textwidth}
         \centering
         \includegraphics[width=\textwidth]{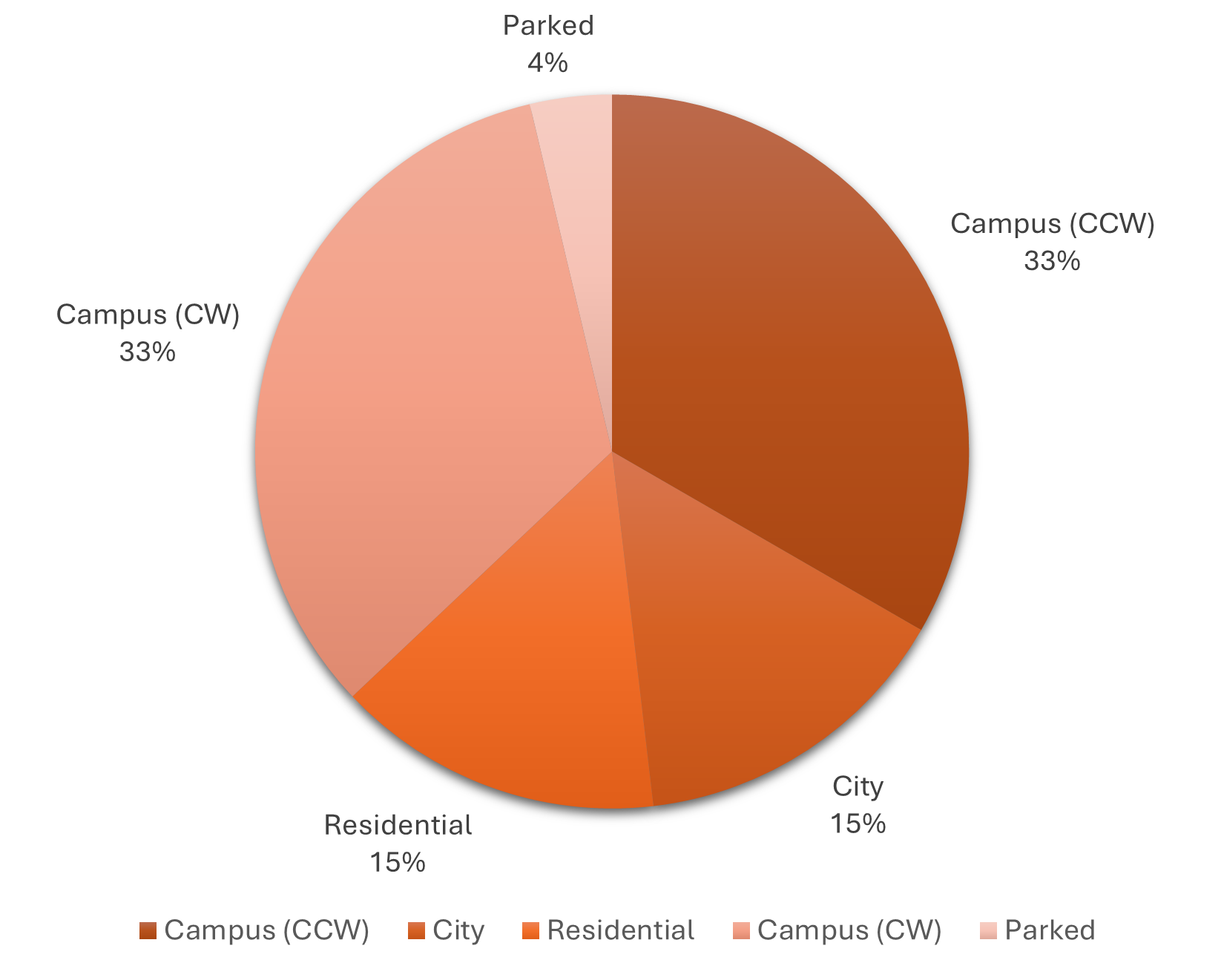}
         \caption{Location}
         \label{fig:pie_c}
     \end{subfigure}
     
     \bigskip
     
     \hspace{40 pt}
     \begin{subfigure}[b]{0.5\textwidth}
         \centering
         \includegraphics[width=\textwidth]{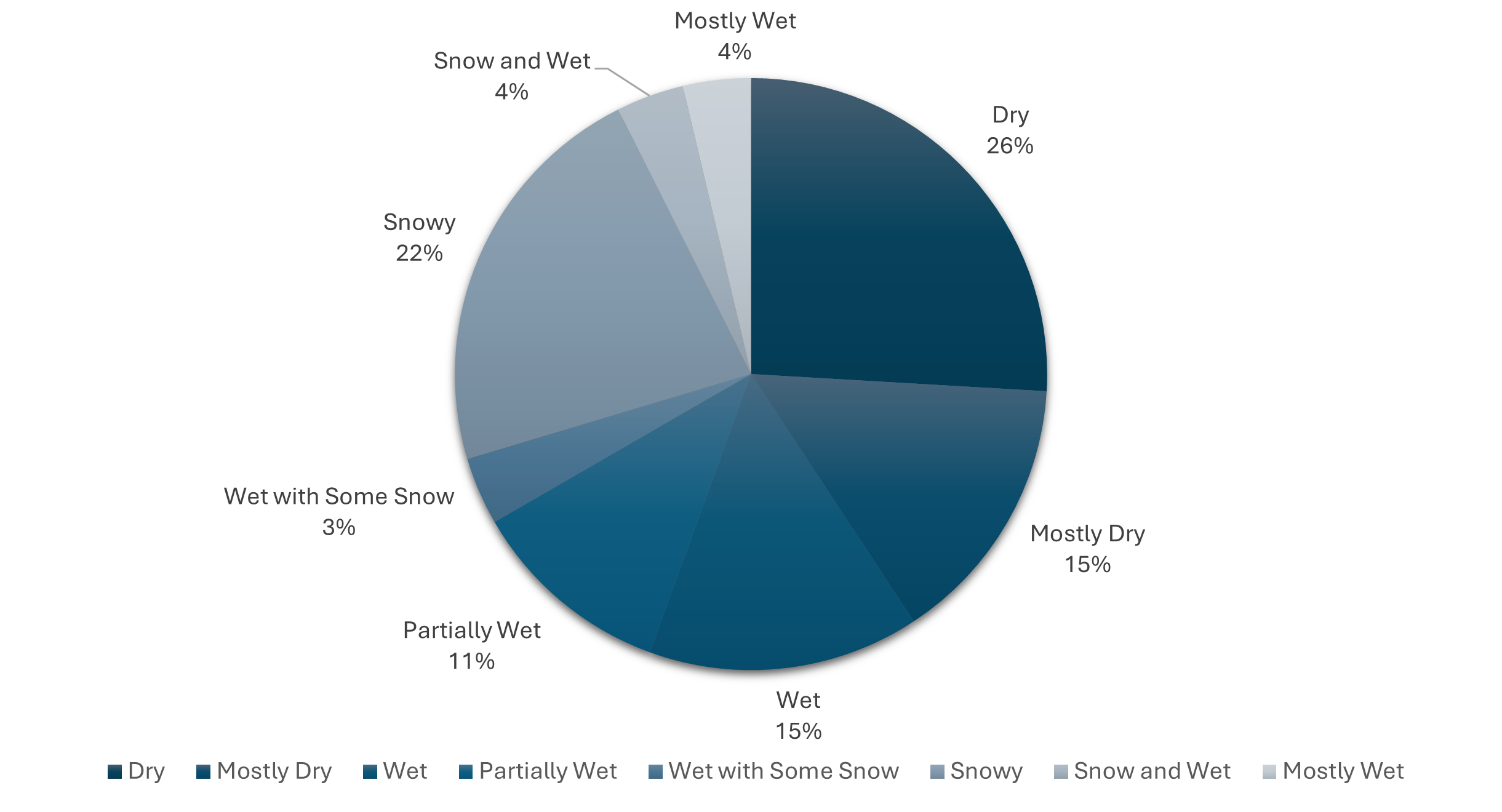}
         \caption{Road Condition}
         \label{fig:pie_d}
     \end{subfigure}
     \begin{subfigure}[b]{0.33\textwidth}
         \centering
         \includegraphics[width=\textwidth]{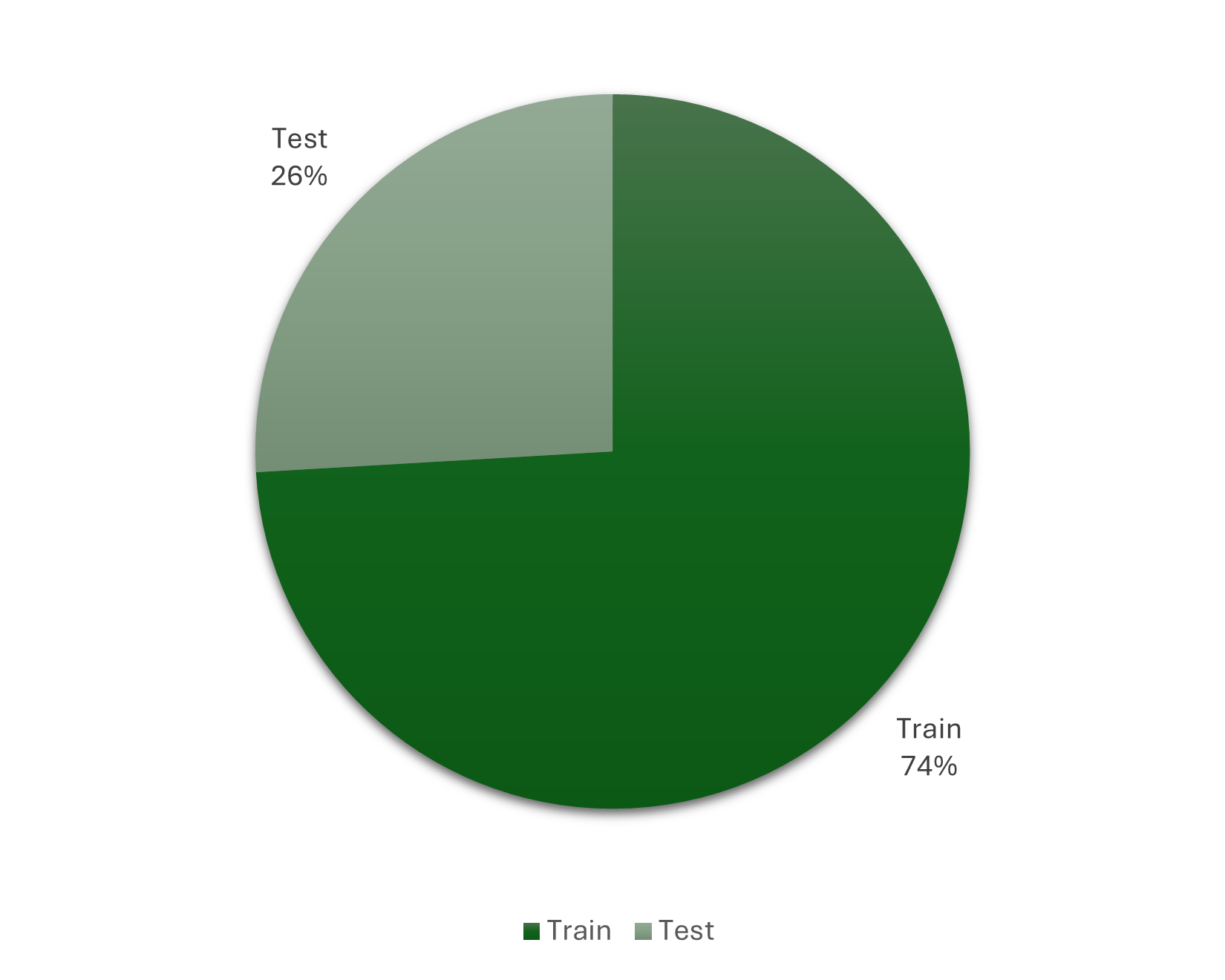}
         \caption{Set}
         \label{fig:pie_e}
     \end{subfigure}
     
    \centering
    \caption{SID sequence distribution across various metrics.}
    \label{fig:pi-charts}
\end{figure*}

\section{Dataset Description and Analysis}
\subsection{Dataset Description}

\subsubsection{Structure and Decomposition}
The dataset is organized into directories based on the different weather classifications. These directories contain subdirectories that represent the individual sequences in the dataset. Each sequence directory follows a specific naming convention:
\begin{equation*}
    \texttt{Location\_Weather\_TimeOfDay\_SequenceNo}
\end{equation*}
In each sequence directory, there are two subdirectories for images captured by the left and right cameras of the stereo-vision camera system. Each image frame is labeled as:
\begin{equation*}
    \texttt{SequenceNo\_Camera\{L/R\}\_FrameNumber.png}
\end{equation*}

\subsubsection{Data Annotation}
The data annotation process went through various stages. Initially, the location, weather conditions, and time of day were noted before capturing each sequence, and then later verified during post-processing. Subsequently, road conditions were manually annotated and categorized for each sequence during post-processing.

\subsection{Statistical Breakdown and Data Insights}

\begin{table*}[t]
\centering
\caption{Detailed breakdown of SID's 27 sequences highlighting the diversity of environmental conditions.}\label{tab:dataset_breakdown}
\resizebox{\textwidth}{!}{
\begin{tabular}{c c c c c c c c c c}
\toprule
\textbf{Sequence No.} & \textbf{Weather Condition} & \textbf{Time of Day} & \textbf{Location} & \textbf{Road Condition} & \textbf{No. of Image Pairs} & \textbf{Total Images} & \textbf{Duration (minutes)} &  \textbf{Set}
&\textbf{Notes} \\
\midrule
1 & Clear & Night & Campus (CCW) & Dry & 8485 & 16970 & 7.1 &   Train
&\\
2 & Clear & Day & Campus (CCW) & Dry & 9005 & 18010 & 7.5 &   Test
&\\
3 & Clear & Night & City & Dry & 5780 & 11560 & 4.8 &   Train
&\\
4 & Clear & Night & City & Dry & 4869 & 9738 & 4.1 &   Train
&\\
5 & Clear & Night & Residential & Dry & 4693 & 9386 & 3.9 &   Test
&\\
6 & Clear & Day & Residential & Dry & 2591 & 5182 & 2.2 &   Train
&\\
7 & Clear & Day & Residential & Dry & 3645 & 7290 & 3.0 &   Train
&\\
8 & Cloudy & Day & Campus (CCW) & Mostly Dry & 7221 & 14442 & 6.0 &   Train
&\\
9 & Cloudy & Day & Campus (CW) & Mostly Dry & 7140 & 14280 & 6.0 &   Train
&\\
10 & Overcast & Day & Campus (CCW) & Wet & 7135 & 14270 & 5.9 &   Train
&\\
11 & Overcast & Day & Campus (CW) & Partially Wet & 7510 & 15020 & 6.3 &   Test
&\\
12 & Rain & Day & Campus (CW) & Wet & 7666 & 15332 & 6.4 &   Test & Minor fog present\\
13 & Rain & Day & City & Partially Wet & 8868 & 17736 & 7.4 &  Train
&BW \\
14 & Rain & Day & City & Wet & 8332 & 16664 & 6.9 &  Train
&BW \\
15 & Rain & Day & Parked & Mostly Dry & 1096 & 2192 & 0.9 &  Train
&BW \\
16 & Rain & Day & Residential & Mostly Dry & 4296 & 8592 & 3.6 &  Train
&BW \\
17 & Snow & Dusk & Campus (CCW) & Wet with Some Snow
& 9887 & 19774 & 8.2 &   Train
&\\
18 & Snow & Day & Campus (CCW) & Snowy & 7988 & 15976 & 6.7 &  Train
&PS [0510], FS [3000] \\
19 & Snow & Day & Campus (CCW) & Snow and Wet & 6934 & 13868 & 5.8 &  Train
&PS [0420], FS [4500] \\
20 & Snow & Night & Campus (CCW) & Snowy & 6883 & 13766 & 5.7 &   Test
&\\
21 & Snow & Day & Campus (CCW) & Mostly Wet & 7038 & 14076 & 5.9 &   Train
&\\
22& Snow& Night& Campus (CW)& Partially Wet& 6638& 13276& 5.5&   Test&
\\
23& Snow& Night& Campus (CW)& Snowy& 6672& 13344& 5.6&   Train&
\\
24& Snow& Dusk& Campus (CW)& Snowy& 7279& 14558& 6.1&   Test&
\\
25& Snow& Day& Campus (CW)& Snowy& 7161& 14322& 6.0&   Train&PS [0180], MS [4000] 
\\
26& Snow& Day& Campus (CW)& Snowy& 6941& 13882& 5.8&  Train&PS [0000], FS [0900] 
\\
27& Snow& Day& Campus (CW)& Wet& 6896& 13792& 5.7&  Train&
\\
\cmidrule{1-10}
&  &  &  &  \textbf{Total} & 178649& 357298& 148.9
&   &\\
&  &  &  &  \textbf{Average}& 6616.6& 13233.3& 5.5
&   &\\
\bottomrule
\end{tabular}
}
\tiny{

} 
\scriptsize{\textit{BW = Below Windshield, PS = Partially Soiled, MS = Mostly Soiled, FS = Fully Soiled, [number] = Frame number for condition}}
\end{table*}

\begin{figure}[t]
     \centering
     \begin{subfigure}[b]{0.24\textwidth}
         \centering
         \frame{\includegraphics[width=\textwidth]{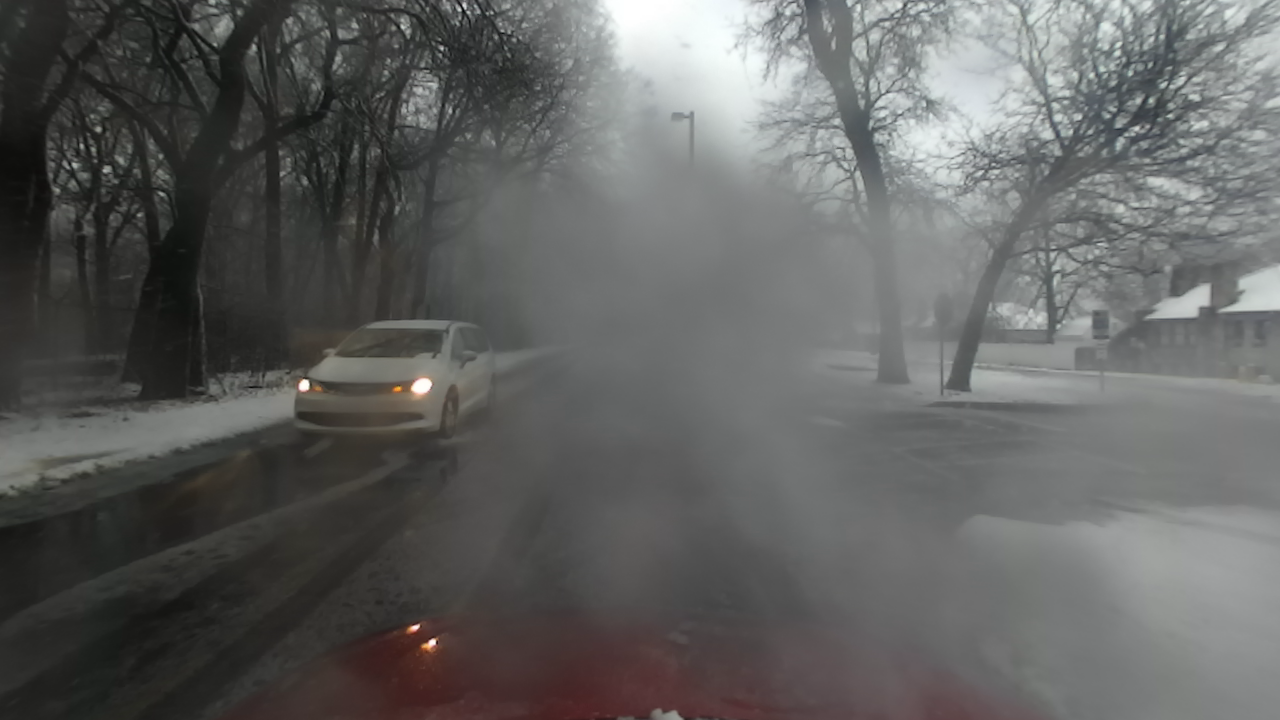}}
         \caption{}
         \label{fig:PC_Left}
     \end{subfigure}
     \hfill
     \begin{subfigure}[b]{0.24\textwidth}
         \centering
         \frame{\includegraphics[width=\textwidth]{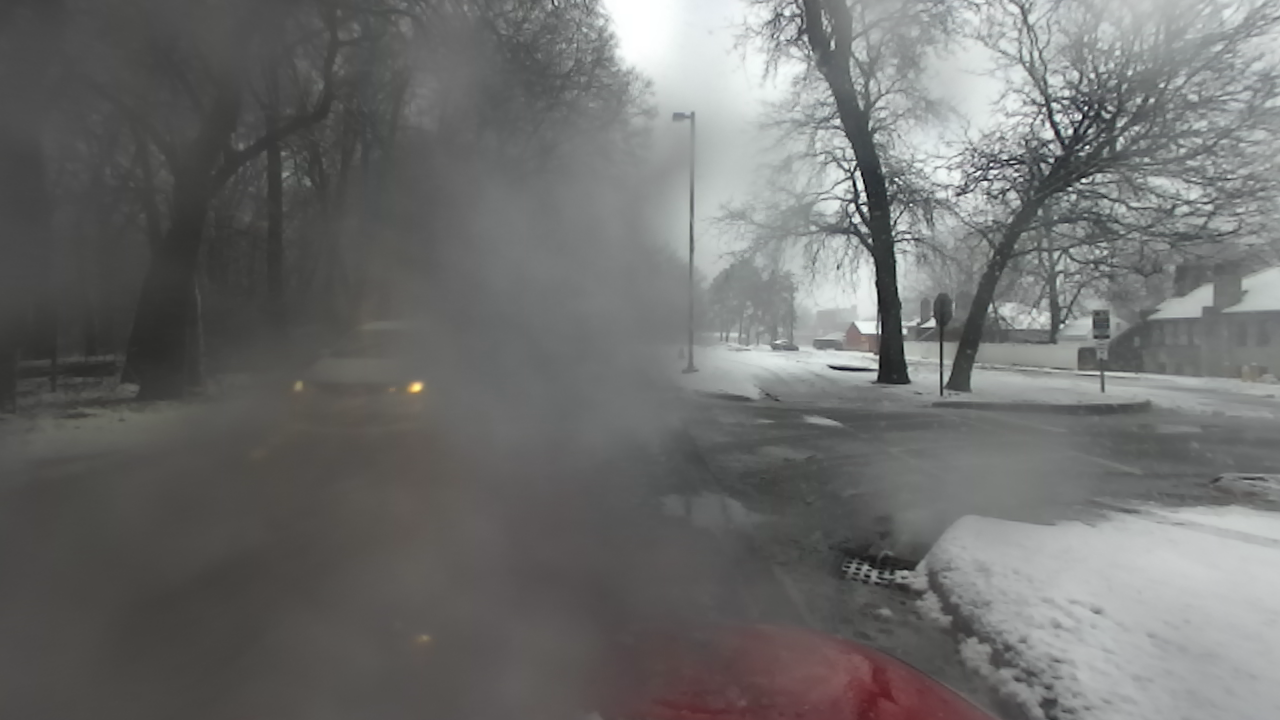}}
         \caption{}
         \label{fig:PC_Right}
     \end{subfigure}
      \caption{A stereo image of the camera with lenses partially soiled by snow, represented by (a) the left and (b) the right images. This stereo image is extracted from sequence no. 26.}
      \label{fig:PC_camera} 
\end{figure}

SID was collected in various weather conditions, including Clear, Cloudy, Overcast, Rain, and Snow. The diverse weather conditions and atmospheric temperatures have led to a varied set of road conditions ranging from Dry to Snowy. Additionally, the data was collected under different lighting conditions and times of day, including Day, Night, and a few at Dusk. 
The overall sequence distribution of our dataset by category is depicted in Fig. \ref{fig:pi-charts}.

As depicted in Fig. \ref{fig:pie_a}, over a third of the dataset was captured in snow conditions. This underscores the importance and uniqueness of this dataset as snow conditions are generally one of the most challenging environments for AVs and ADAS systems. This is complemented by the challenging road conditions such as Snowy and Wet constituting a significant portion of this dataset, as highlighted in Fig. \ref{fig:pie_d}. Road conditions can significantly impact the performance of computer vision models, particularly those related to autonomous driving, road condition monitoring, and vehicle navigation. Furthermore, as shown in Fig. \ref{fig:pie_b}, over a quarter of the dataset's sequences are collected during nighttime, resulting in various challenging low-light conditions that require robust perception algorithms to handle them. Including diverse training data, such as varied road and lighting conditions, ensures that the developed models are robust and can handle different scenarios. This is particularly crucial for applications like AVs, which need to operate safely in all conditions.

Regarding the location, the pre-planned campus track comprises the majority of the recorded sequences split between a clockwise (CW) and a counterclockwise (CCW) track, as shown in Fig. \ref{fig:pie_c}. These sequences can be utilized to evaluate the effect of various environmental conditions on critical perception algorithms such as visual odometry.

Finally, to support benchmarking for model development and evaluation, our dataset is partitioned into training and testing sets. We propose a 74--26\% train-test split, resulting in 20 sequences for training and 7 for testing, as depicted in Fig. \ref{fig:pie_e}. This allocation aims for a balance across all dataset dimensions, offering ample data for model training while establishing a robust benchmark for testing.

Table \ref{tab:dataset_breakdown} outlines the dataset distribution across 27 sequences, detailing the diverse conditions and corresponding data sampled. The table specifies parameters including weather condition, time of day, and location, further noting specifics like road conditions and the number of image pairs captured.

The composition of the dataset demonstrated in Table \ref{tab:dataset_breakdown} ensures a comprehensive representation of varying environments, enabling the evaluation of perception systems across a spectrum of operational challenges. SID includes scenarios with challenging conditions such as intense snowfall, which may lead to camera lens soiling, thereby adding complexity to the scenarios presented. These scenarios, exemplified in Fig. \ref{fig:PC_camera}, are crucial for developing robust perception algorithms.

In total, SID comprises over 178k stereo image pairs (357k individual images) collected over approximately 149 minutes. The presence of weather-related adverse effects, like camera lens soiling documented in specific sequences, emphasizes the dataset's value for testing the limits of autonomous perception systems. Such instances are critical for assessing the performance of technologies aimed at real-world applications, where unpredictability is a constant factor.

\begin{figure*}[t]
    \centering
    \begin{subfigure}[b]{0.243\textwidth}
        \frame{\includegraphics[width=\textwidth]{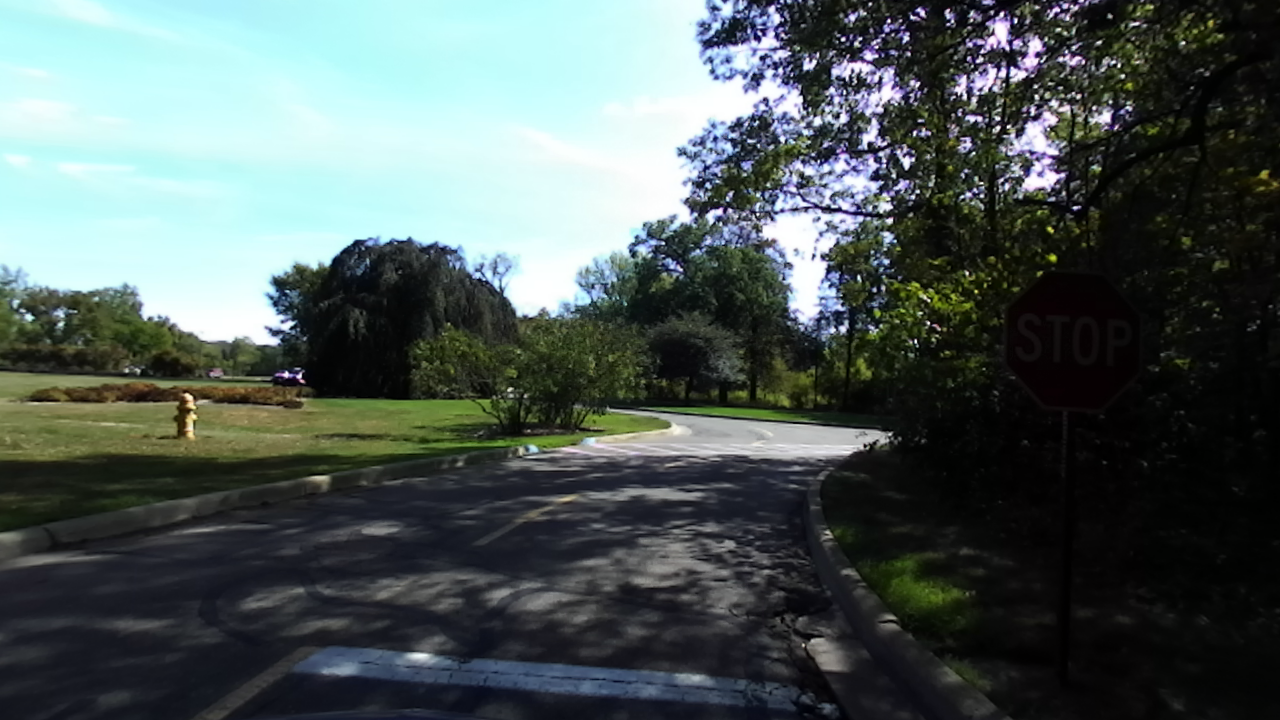}}
        \caption{High dynamic range scene}
    \end{subfigure}        
    \hfill
    \begin{subfigure}[b]{0.243\textwidth}
        \frame{\includegraphics[width=\textwidth]{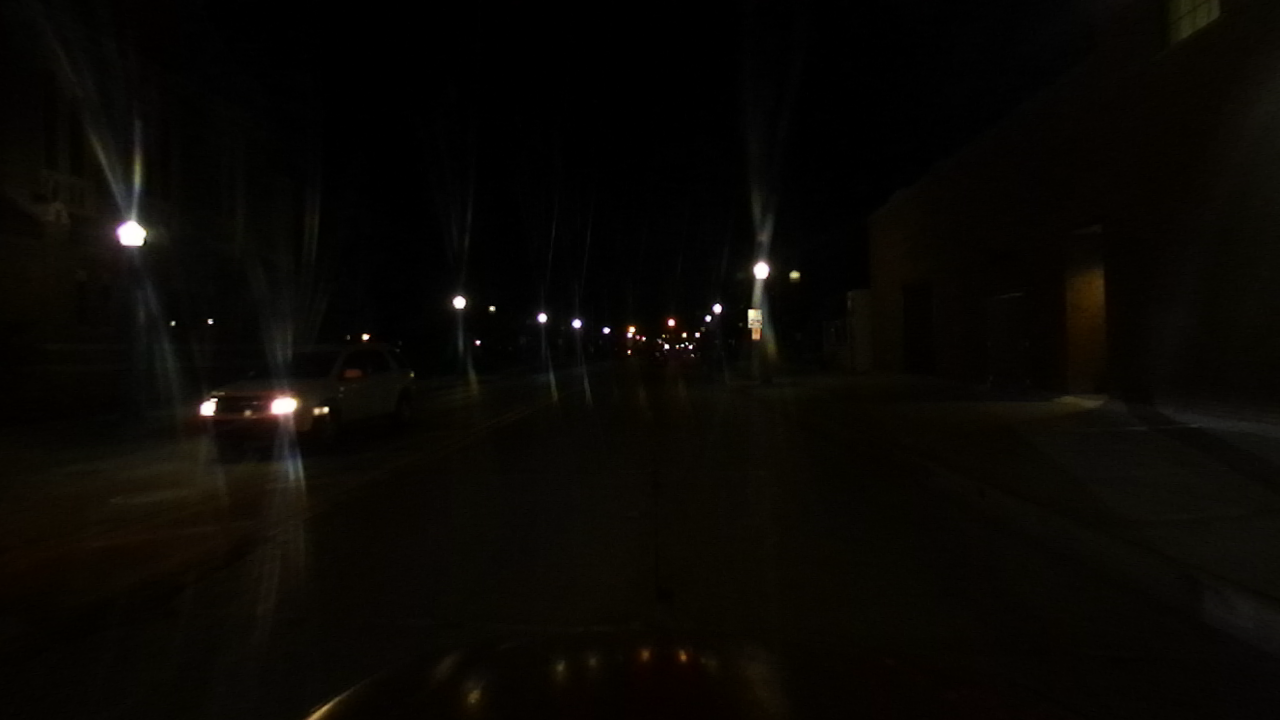}}
        \caption{Low-light environment}
    \end{subfigure}    
    \hfill
    \begin{subfigure}[b]{0.243\textwidth}
        \frame{\includegraphics[width=\textwidth]{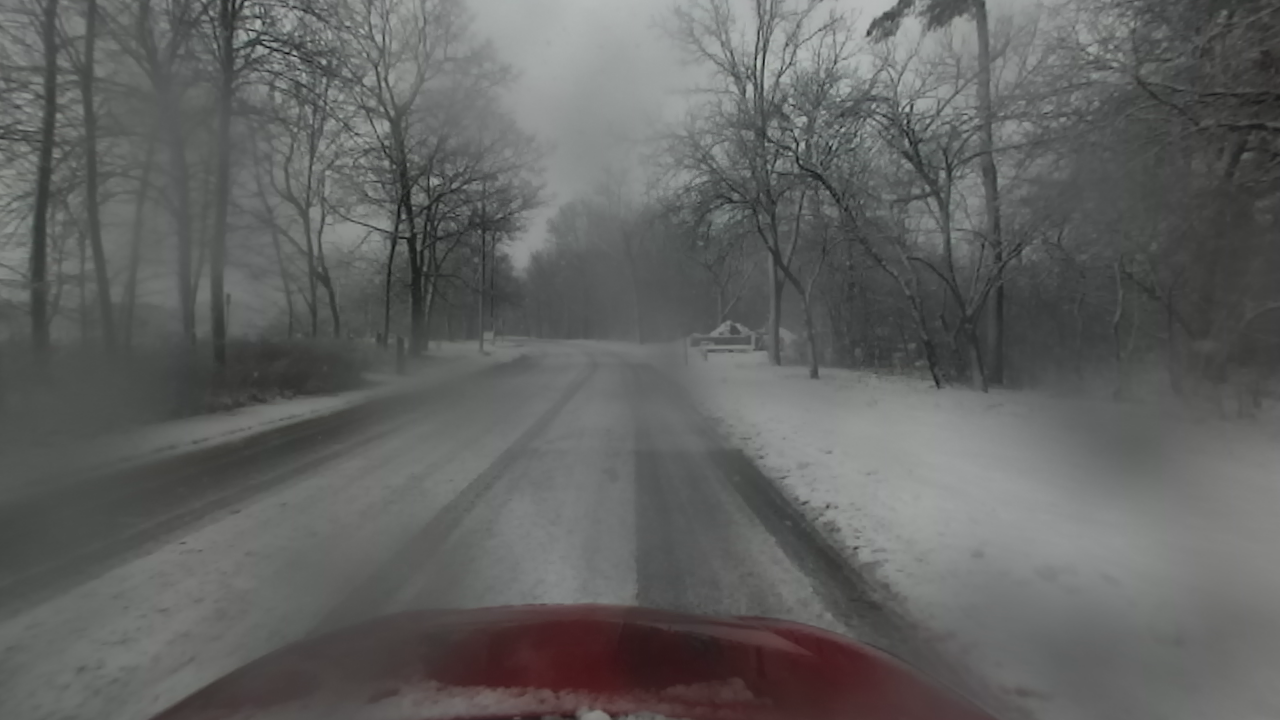}}
        \caption{Soiled lens}
    \end{subfigure}
    \hfill
    \begin{subfigure}[b]{0.243\textwidth}
        \frame{\includegraphics[width=\textwidth]{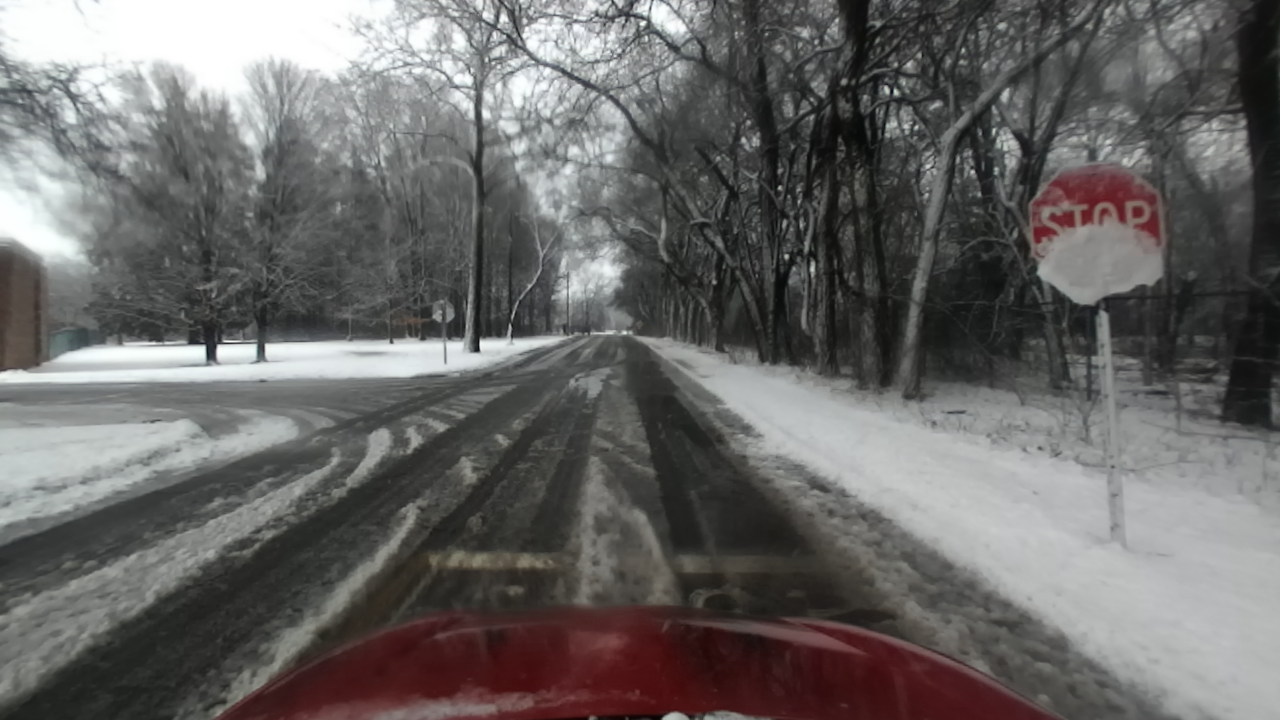}}
        \caption{Snow-covered traffic sign}
    \end{subfigure}
    \caption{Challenging scenarios represented in SID: (a) high dynamic range scenes, (b) low-light environments, (c) camera lens partially soiled by water droplets, and (d) snow-covered traffic signs.}
    \label{fig:challenging_scenarios}
\end{figure*}

Fig. \ref{fig:challenging_scenarios} highlights some of the particularly demanding scenarios present in the dataset, including high dynamic range scenes and low-light environments. These situations emphasize the potential of the dataset for advancing research in fields where visual perception is the primary input source.

\section{Applications and Use-Cases}

The versatility of SID—encompassing varied weather, lighting conditions, and environmental settings—enables its utility across several domains. Below, we explore a selection of prominent applications and potential use cases that could significantly benefit from our dataset:

\subsection{Weather Classification and Understanding} SID is designed to support weather classification algorithms, which are vital for both environmental monitoring and the operational reliability of autonomous navigation systems \cite{lu2014two}. Beyond the weather labels, SID provides sequence-level annotations including time of day and road conditions. Such comprehensive metadata enables the development of advanced models that can classify and react to varying deriving conditions in real time, promoting the reliability of AVs and ADAS. Leveraging deep learning, researchers can utilize SID to improve the robustness and accuracy of classification systems, equipping them to handle the complexities of varied and challenging environmental conditions \cite{kang2018deep}.

\subsection{Robust Stereo Vision Algorithms} Stereo vision is essential for depth estimation—a key component in both robotics and autonomous navigation systems. Given that our dataset includes imagery from a vast range of adverse weather conditions, it provides an excellent opportunity for research into enhancing the resilience of stereo vision algorithms \cite{wang2007brdf, maddern2020real}. It can aid in improving stereo matching in difficult visual scenarios, such as lens soiling caused by rain, snow, or fog, as demonstrated in Fig. \ref{fig:PC_camera}.

\subsection{Day/Night Image Enhancement} Our dataset's broad range of light conditions is ideal for furthering research into image enhancement, particularly for applications that require robust visibility from day to night. Enhancement techniques that optimize dynamic light adaptation are critical for driver-assistance systems and AVs, where changing light conditions can dramatically affect visibility \cite{pizer1987adaptive}. Innovations in this space, particularly those using deep learning approaches, offer promising solutions for improving low-light image clarity and detection efficacy \cite{jiang2021enlightengan}.

\subsection{Autonomous Vehicle Navigation and Control Systems} The various driving conditions captured in SID can help refine the navigation and control algorithms of AVs. Simulating exposure to challenging real-world situations, such as those depicted in Fig. \ref{fig:challenging_scenarios}, provides invaluable training data for vehicles to deal with reflections, glare, and reduced visibility, which are common in difficult weather conditions \cite{montemerlo2008junior, chen2015deepdriving}.

\subsection{Obstacle and Hazard Detection} The ability to detect hazards and obstacles is critical to the safety and reliability of autonomous systems \cite{van2018autonomous}. Our dataset's rich variety of environmental settings is beneficial for the development of more robust detection algorithms that can account for the impact of weather conditions on sensor information. Systems developed with this data could greatly improve the detection reliability under diverse operational states which is an aspect of research essentially underscored by the significant representation of snowy conditions in our sequences.

\section{Conclusion}


In this paper, we have introduced SID, a comprehensive stereo-image dataset designed to bridge the gap in autonomous driving research under adverse weather and lighting conditions. Captured using the ZED stereo camera, SID encompasses over 178k pairs of images with detailed sequence-level annotation of weather, lighting, and time of day across various locations. The dataset’s robust variety, encompassing challenging conditions such as heavy snowfall, rain, and poor lighting, makes it a critical resource for driving advancements in weather classification, perception algorithms, image processing, and the broader domain of autonomous navigation.

Notably, SID is not is not without limitations. Currently, it lacks extensive object detection labels (including bounding boxes and corresponding classifications) for the numerous relevant objects such as pedestrians, vehicles, and traffic signs that it contains. The dataset's rich background makes it an excellent candidate for further annotation efforts. Subsequent work could leverage the existing depth cues and challenging scenarios to create a benchmark for object detection under various environmental conditions that stand to advance the field of autonomous systems.

\bibliographystyle{IEEEtran}
\bibliography{references}
\vspace{12pt}
\end{document}